\documentclass[11pt,a4paper]{article}
\usepackage[hyperref]{acl2020}
\usepackage{times}
\usepackage{latexsym}

\usepackage{microtype}

\usepackage{amsmath}
\usepackage{array}
\usepackage{graphicx}
\usepackage{wrapfig}

\aclfinalcopy 

\title{Situated Multimodal Control of a Mobile Robot: \\ Navigation through a Virtual Environment} 

\author{Katherine Krajovic, Nikhil Krishnaswamy, Nathaniel J.
Dimick, \\ {\bf R. Pito Salas, and James Pustejovsky} \\
  Brandeis University Department of Computer Science\\
  Waltham, MA \\
  \texttt{\small\{krajovic,nkrishna,ndimick,pitosalas,jamesp\}@brandeis.edu}}

\date{}

\begin{document}
\maketitle
\begin{abstract}
\vspace{-2mm}
We present a new interface for controlling a navigation robot in novel environments using coordinated gesture and language.  We use a TurtleBot3 robot with a LIDAR and a camera, an embodied simulation of what the robot has encountered while exploring, and a cross-platform bridge facilitating generic communication.  A human partner can deliver instructions to the robot using spoken English and gestures relative to the simulated environment, to guide the robot through navigation tasks.

\end{abstract}

\vspace{-4mm}
\section{Introduction}
\label{sec:intro}
\vspace{-2mm}

Recent developments in HRI have given rise to new capabilities in how we communicate with and control robotic devices, including
navigational robots \citep{tellex2020robots}. Key to this change is the emergence of language interfaces, which have allowed for more familiar and natural interactions with robots through language in limited dialogue settings \citep{roque2020developing}.  Another communicative avenue that has recently emerged is the use of gesture  \citep{williams2019mixed}. What is still lacking, however, is the ability to interact with a robotic agent multimodally, with integrated language and gesture in a dynamic communication. 

Here, we describe a human-robot interaction through the integration of multiple modalities in VoxWorld, a platform for modeling and communicating with computational agents, which contains a simulation environment and a semantic model interpreted with the language VoxML \citep{pustejovsky2016LREC}. Simulation plays a critical role in communication between humans and robots by creating a shared common ground (epistemic model) of the co-inhabited environment. The simulation demonstrates the knowledge held by the robot publicly, which is needed to ensure shared understanding with the humans in the activity.

In our system, a human user and a robot exist in a co-situated space that is mediated by a virtual environment displayed on a screen, such that the human can see a virtual rendition of the environment the robot has explored, and of the robot's current perspective view.  The human can then gesture to objects and locations on the screen, either in a perspective or omniscient view, and speak about them in English, e.g., ``{\it go there},'' ``{\it go to that wastebasket and turn around},'' or ``{\it find the blue block}.''  Deictic gestures are grounded to coordinates on the screen which are transformed to equivalent coordinates in the robot's ROS environment, allowing the robot to execute native navigation commands, e.g, $go\_to(x,y)$.  The robot can likewise communicate status updates back to the human which are then spoken out through text-to-speech.

\section{VoxWorld Platform}
\label{sec:voxworld}
\vspace{-2mm}

{\it VoxWorld} is the simulated situated grounding platform that facilitates communication between humans and intelligent agents, here a robot.  It is built on VoxML, which encodes object, event, and relation semantics exploitable by computational reasoners in simulated environments.

The Unity game engine-based software VoxSim \cite{krishnaswamy2016voxsim} visualizes events from VoxML encodings, and provides the virtual environment in which VoxWorld agents perceive and reason about their surroundings.

Situated environments in the VoxWorld platform provide agents a dynamic point-of-view in a virtual or simulated world.  The virtual worlds assume an embodiment of the agent in the environment, and VoxWorld provides a platform for the agent to express its interpretation of the world via its {\it embodied theory of mind} \citep{johnson2002conditionals}.  The agents present semantic interpretation of linguistic expressions, object perception, and environmental awareness through a simulation \citep{pustejovsky2019situational}.

VoxSim (and VoxWorld) contains a neural model of underspecified motion verb parameters \cite{krishnaswamy2017montecarlo} that assigns them appropriate values to operationalize predicates in terms of primitives like ``translate'' and ``rotate,'' the same primitives executable by a real robotic agent (\S~\ref{ssec:specs}).

\vspace{-2mm}
\section{Robotic Agent}
\label{sec:robot}
\vspace{-2mm}

The robotic agent we have developed, dubbed ``Kirby,'' uses a Turtlebot3 from Robotis, with differential drive, non-holonomic steering, further equipped with a LIDAR (laser-based
distance sensor) and a color camera. The robot has two compute boards: an Arduino-based OpenCR controlling the motors, and a Raspberry Pi running Linux and ROS. ROS, the Robot Operating System, is a distributed operating system widely used in research and industry. The robot operates on a flat floor with walls forming corridors, corners, openings, etc. with Aruco fiducials denoting known physical objects.

\begin{wrapfigure}{l}{0.12\textwidth}
\vspace{-6mm}
\begin{center}
\includegraphics[width=0.12\textwidth]{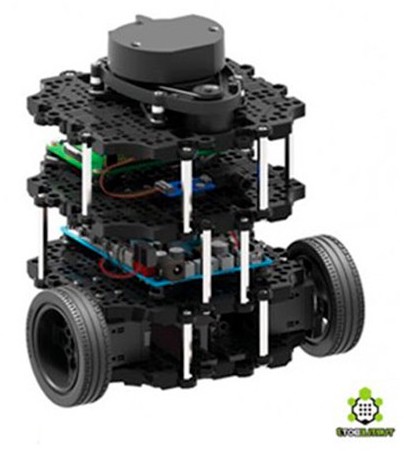}
\vspace{-4mm}
\caption{TurtleBot3.}
\label{fig:tb3}
\end{center}
\vspace{-6mm}
\end{wrapfigure}

Our lab was suddenly closed due to the COVID-19 pandemic, restricting access to physical robots, which led us to a fully simulated scenario, using the the Gazebo simulation platform, a well established and powerful 3D simulation environment.  Gazebo provides both the simulation of the physical surroundings of floor plane, walls, corridors, 3D space, and simple physics (mass, weight, friction, gravity), as well as the simulation of the TurtleBot3 Robot itself, including physical characteristics (e.g., shape, mass) and dynamic simulation of its motors, wheels, sensors and computation. The simulated robot will respond to precisely the same commands as the physical robot. There is a degree of randomness in the physical world (e.g. a bump in the carpet or change in lighting) which is not fully simulated.

\vspace{-2mm}
\subsection {Specifications, Abilities, Parameters}
\label{ssec:specs}
\vspace{-2mm}

Kirby supports several basic movement commands, including: {\it go forward}, {\it go to}, {\it turn left}, {\it turn right}, and {\it patrol}. A ``go forward'' command takes an optional parameter $x$ (default 1m) and moves Kirby $x$ meters ahead of its current location.  ``Go to'' takes $x$ and $y$ and moves Kirby to the specified $(x,y)$ coordinate on its map. To execute both of these, Kirby will navigate around any obstacles that it encounters. A ``turn left'' or ``turn right'' command takes an optional parameter $d$ (default 90$^{\circ}$), and turns Kirby $d^{\circ}$ counter-clockwise or clockwise respectively. 

``Patrol'' takes optional parameters $s$, $r$, $i$.  Kirby explores its environment in concentric $s$-sided polygons with vertices $r$ meters from the origin, and vertices of each subsequent polygon $i$ meters greater than the previous; defaults are 16 sides, initial radius of 1.5m and increment of 1.0m.

Kirby can take in multiple commands and will execute them in order. It also supports flow of control commands {\it stop}, {\it continue}, {\it cancel}/{\it cancel all}, and {\it go back}. ``Stop'' pauses the movement Kirby is in the process of executing, while ``continue'' unpauses it. ``Cancel'' deletes the current movement from its execution queue, and ``cancel all'' deletes all movements in the queue. ``Go back'' deletes all queued movements and also sends Kirby back to the location it was in before the current movement.

\vspace{-2mm}
\subsection{Environment}
\label{ssec:env}

The environment Kirby operates in consists of three major elements: Robots (either physical or simulated), the {\it Robotic Services Bridge} (RSB) and the Unity-based VoxSim environment described above.  These descriptions of the environment apply to both the physical robot and the simulated robot.

\vspace{-2mm}
\subsubsection{Fiducials}
\label{sssec:fiducials}
Fiducial markers denote specific points and directions in space. Typically in robotics they are square black and white signs, approximately 10 $\times$ 10cm, akin to QR Codes. 
\begin{wrapfigure}{l}{0.22\textwidth}
\vspace{-6mm}
\begin{center}
\includegraphics[scale=0.17]{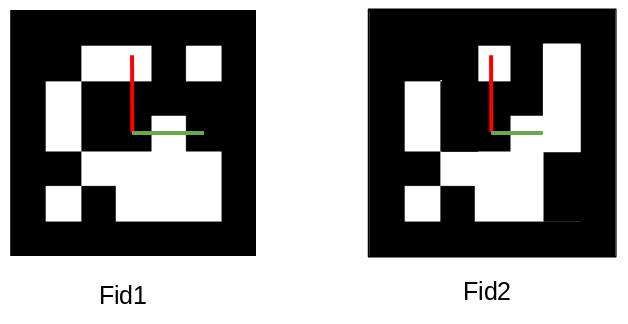}
\vspace{-6mm}
\caption{Two fiducials.}
\label{fig:fiducials}
\end{center}
\vspace{-6mm}
\end{wrapfigure}

With a camera and appropriate software, the robot can detect fiducials, and specifically their distance and orientation relative to the camera (using the known size of the fiducial and perspective distortion.) The fiducial can also encode a numeric identifier. When the robot's camera first sees a fiducial, its location, orientation and identifier are reported through RSB.

\vspace{-2mm}
\subsubsection{Robotic Services Bridge}
\label{sssec:rsb}
RSB is designed to enable control and supervision of robots to other systems in a flexible, platform-independent way.  RSB maintains a distributed and shared key-value store accessible by any client such as Kirby. All communication, commands and monitoring of the robotic side is done by reading and writing keys (see Table~\ref{tab:keys}) in this shared store (implemented as a Redis cache. cf. \citep{tzafestas1991blackboard}).
\begin{table}
\begin{center}
\small
    \begin{tabular}{ | p{4.5em} | p{17em} | }
    \hline
    {\bf Key} & {\bf Function} \\
    \hline
    {\sc Map} & Obstacles (walls) the robot has encountered, represented as line segments computed from LIDAR data using a line segment merging method \citep{tavares1995new}. \\
    \hline
    {\sc Odom} & Current location, direction, forward and rotational velocity of the robot. \\
    \hline
    {\sc Kirby} & Commands from VoxWorld to Kirby (see \S~\ref{ssec:communication}). \\
    \hline
    {\sc Fiducials}  & List of detected fiducials, including identifier and 3D location/orientation. \\
    \hline
    {\sc Kirby\_} & Stream of status information for \\
    {\sc Feedback} & communicating to the user and troubleshooting. \\
    \hline
    {\sc Bridge\_} & Commands to reset the RSB. \\
    {\sc Reset} & \\
    \hline
    \end{tabular}
\end{center}
\vspace{-4mm}
\caption{Keys of the Robotic Services Bridge.}
\label{tab:keys}
\vspace{-8mm}
\end{table}

\vspace{-2mm}
\subsection{Communication and Control}
\label{ssec:communication}
\vspace{-2mm}

\begin{figure}[h!]
    \centering
    \includegraphics[height=1.2in]{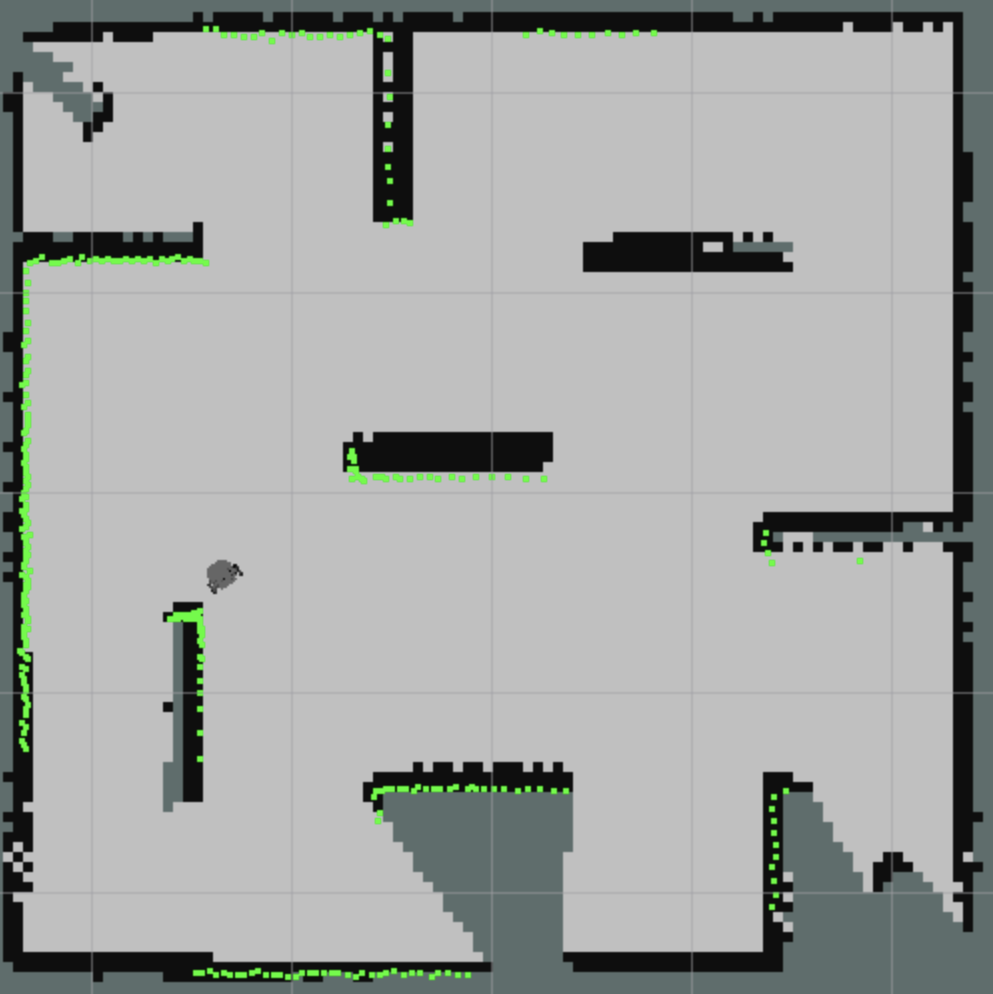}
    \includegraphics[height=1.2in]{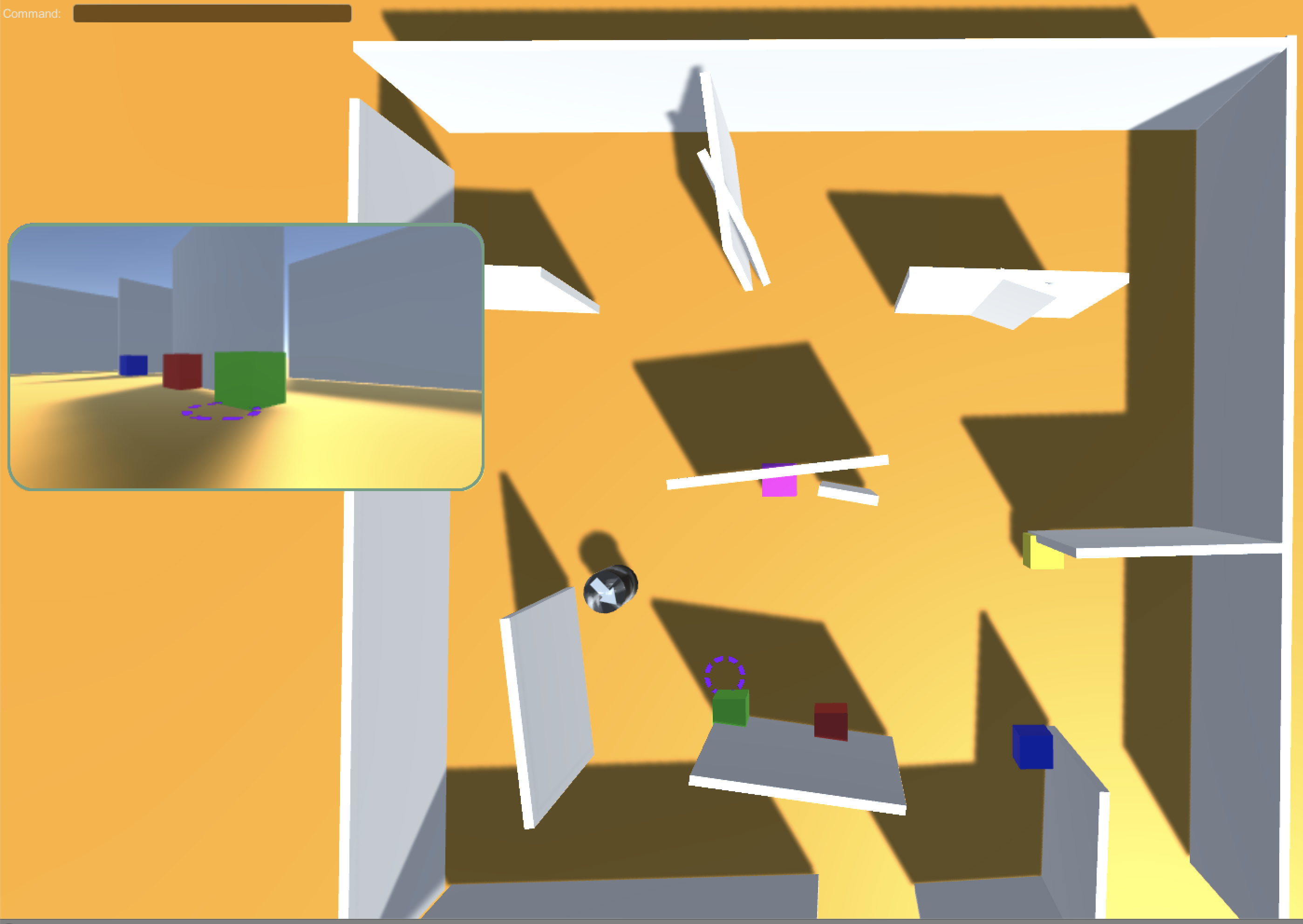}
    \caption{Visualized LIDAR data (L) and Kirby's VoxWorld interpretation (R). The main view shows an omniscient view and the inset shows Kirby's perspective.}
    \label{fig:screenshot}
\vspace{-4mm}
\end{figure}

As Kirby navigates its environment, line segments generated from the LIDAR data as well as position and speed updates are posted to the {\sc Map} and {\sc Odom} channels of the RSB.  Fiducials encountered are posted to the {\sc Fiducials} channel.  Kirby's VoxWorld environment listens for updates on these channels and builds a simulated visualization of what Kirby encounters for its human partner to see and interpret (Fig.~\ref{fig:screenshot}).  The virtual environment presents the interpreted realspace data from Kirby's perspective and from an omniscient bird's-eye view.  Kirby's perspective updates as Kirby moves through the environment, and the human can navigate the bird's-eye view using mouse and keyboard, but also real-time recognized gesture, e.g, {\it push left}/{\it right} to switch camera views, or {\it servo left}/{\it right}/{\it back} to rotate the camera (see \S~\ref{sssec:multimodal} for the default gesture list).

At startup, Kirby will provide a map of the world from its initial perspective (\S~\ref{sssec:mapml}).  The user can give spoken commands and gesture to regions or fiducials, which are currently used as proxies for objects of interest, in the VoxWorld rendering.  Interpretations of these instructions are then communicated back to Kirby through the RSB's {\sc Kirby} channel (\S~\ref{sssec:multimodal}).

As Kirby executes instructions and uncovers larger portions of the world, the map, as well as the position and orientation of Kirby's ``avatar'' in VoxWorld will update over time. Kirby provides feedback codes (published on the RSB's {\sc Kirby\_Feedback} channel) to indicate status, success, failure, or the need for user input, which are transformed into messages displayed on screen in VoxWorld and spoken aloud using text-to-speech.

\vspace{-2mm}
\subsubsection{Map Interpretation}
\label{sssec:mapml}

2D line segment endpoints ($a_1$,$b_1$,$a_2$,$b_2$) transform from ROS space to Unity as $\lVert(-b_2,a_2)-(-b_1,a_1)\rVert$-meter-long cuboids at XZ-coordinates
$(\frac{(-b_1.x,a_1.x) + (-b_2.x,a_2.x)}{2},\frac{(-b_1.z,a_1.z) + (-b_2.z,a_2.z)}{2})$, rotated $-sin^{-1}(\lvert(-b_2,a_2)-(-b_1,a_1)\rvert.z)\times sgn(\lvert(-b_2,a_2)-(-b_1,a_1)\rvert.x)$ radians around the Y-axis.

\begin{wrapfigure}{r}{.2\textwidth}
\vspace{-6mm}
\begin{center}
    \includegraphics[width=.2\textwidth]{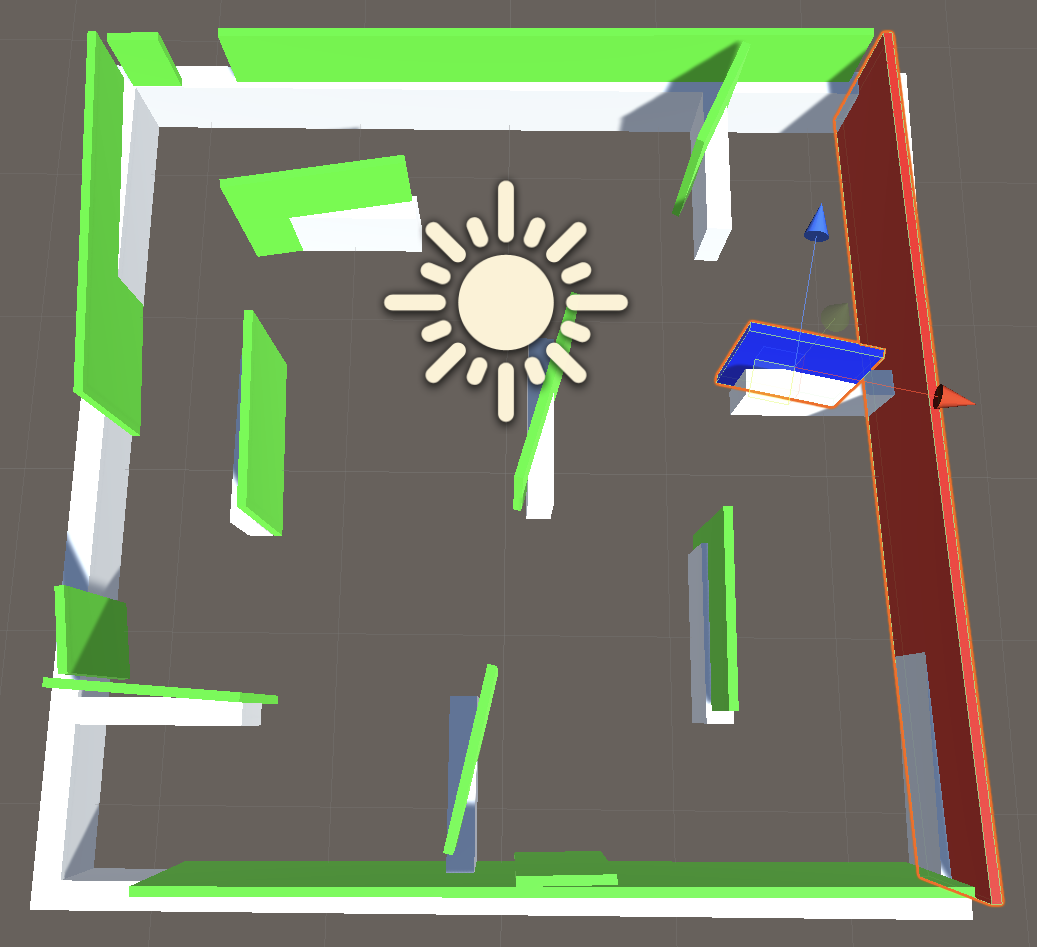}
    \caption{Sample ground truth (white) and annotated LIDAR-derived line segments (color).}
    \label{fig:ml}
\end{center}
\vspace{-10mm}
\end{wrapfigure}

Being extracted from LIDAR data, these imprecise line segments may change, merge, or vanish as Kirby navigates through the world.  This constant variation presents difficulties for the human partner, as a certain consistency is required for an interpretable world.  To combat this, and present an interpretable simulated view to the human partner, we annotate raw LIDAR-derived line segments relative to the ground truth, with an action space of aligning and closing gaps between pairs of line segments (Fig.~\ref{fig:ml}).  We then train a 4-layer deep neural network on this data to determine additional transformations to be made in Unity space.

This presents a more consistent and cohesive interpretation of the world to the human partner, that becomes more accurate as Kirby approaches obstacles.

\vspace{-2mm}
\subsubsection{Speech and Gesture}
\label{sssec:multimodal}

Through VoxSim and the RSB, Kirby consumes instructions from the human partner in spoken English.  These commands can be those enumerated in \S~\ref{ssec:specs}, or instructions like ``go to the first fiducial on the right,'' that are parsed and executed through the VoxSim-based event manager.  Kirby can also ``see'' the human through custom gesture recognition running on deep convolutional neural networks trained over Microsoft Kinect\textsuperscript{TM} data \citep{mcneely2019user}.  By default there is a fixed gesture set that the system responds to (cf.~\citeauthor{li2012hand}~\citeyearpar{li2012hand}), including pointing, {\it thumbs up}/{\it down}, a ``{\it claw},'' {\it push left}/{\it right}, {\it beckon}, and ``{\it servo}'' (an iterated push or ``nudge'') {\it left}/{\it right}.  The system can use random forests to learn new gestures ``live'' (cf.~\citeauthor{matuszek2014learning}~\citeyearpar{matuszek2014learning}), which can be linked with linguistic instructions, e.g., an {\it L} shape for``go forward 1m, turn right, then go forward 2m.''

The user can point to locations or objects in the VoxWorld environment and ground their instructions to those entities, e.g., ``go there,'' ``go to that one,'' ``turn this way,'' ``a little further,'' etc.  This diversity of commands reflects how different modalities ground different types of information (e.g., descriptive language for ``first block on the right'' vs. deictic gesture and/or demonstratives for locations---``there'').  User input is interpreted into sequences of basic commands (\S~\ref{ssec:specs}) that are pushed onto Kirby's eponymous command queue.   If a linguistic instruction fails due to either poor speech recognition or parsing failure, Kirby will either not respond or reply with a message ``I didn't understand'' or similar.  The human can additionally use gesture to communicate if language fails.

Instructions from Kirby's human partner must be situated in the world that Kirby inhabits, including distinct object resolution and specific coordinate localization.  Coordinates are transformed from Unity space to ROS space ($(-b_1,a_1),(-b_2,a_2) \rightarrow (a_1,b_1),(a_2,b_2)$) so the robot can execute commands using native functionality.

While executing, Kirby publishes feedback to the user. It sends an update when beginning or completing a movement, or if it determines that a goal is unachievable (e.g., ``looking for path to ($x$, $y$),'' ``successfully navigated to ($x$, $y$),'' ``unable to complete goal''). It also publishes messages if goals are canceled, paused, or resumed (e.g., ``paused current goal,'' ``canceled [all] goal(s),'' ``restarting current goal,'' ``waiting for commands'').

If, in attempting to complete a movement, Kirby begins navigation but then determines it cannot reach the goal location, it requests input as to whether it should go back to its original location or continue executing from the new location (``moved from expected path and failed to reach goal,'' ``user input is required: keep going OR go back''). 

\vspace{-2mm}
\section{Scenario and Evaluation}
\label{sec:scenario}
\vspace{-2mm}

A typical scenario begins with Kirby entering a new space. The human gives navigation commands such as ``go forward'' or ``patrol'' and as the robot discovers more of the environment, it sends information back to VoxWorld to construct the simulated representation, including fiducials representing notable objects.  In Fig.~\ref{fig:screenshot}, Kirby detects 5 fiducials (visualized as boxes)---3 on the right and 2 on the left.  The human determines a target for Kirby and communicates it in English, e.g., ``go to the first one on the right.''  ``First one on the right'' is situated in VoxWorld and resolved to specific coordinates which are then communicated to Kirby with a {\it go to} instruction.  Kirby, while navigating to the target, encounters an obstacle and has to go around it, perhaps finding along the way that there is no path to the desired location. This is communicated back to the human: ``user input is required: keep going OR go back.''  The human, being able to see the simulation both in the bird's-eye view and from Kirby's perspective, can point to an achievable location near the target object (the purple circle in the inset of Fig.~\ref{fig:screenshot}), and tell Kirby ``go {\it there}.''  The updated coordinates are then sent to the robot.

As this system is very new, evaluation is still being planned.  We plan to give participants a target object to find that they then have to direct Kirby towards, with no prior language coaching and a verbal description of the gestures and evaluate time to completion and usage of various modal techniques.

\vspace{-2mm}
\section{Conclusion and Discussion}
\label{sec:conc}
\vspace{-2mm}

If a robot can receive information from a human collaborator in a linguistic or gestural modality and interpret that relative to its current physical circumstances, it can create an epistemic representation of the information provided by the human. In the absence of any modality of expressing that representation independently, the human cannot verify or query what the robot agent is actually perceiving or how that perception is being interpreted.  A simulation environment, such as VoxWorld presented here, provides a venue for the human and robot to share an epistemic space, and any communicative modality that can be expressed within that space (e.g., linguistic, visual, gestural) enriches the number of ways that a human and a robot can communicate on object and situation-based tasks.

Scenarios such as \S~\ref{sec:scenario} serve as proxies for situations where robots assist humans in environments where humans cannot go safely, e.g., a burning building or constricted space, but must rely on the interpretive capacity and background knowledge of humans to complete their task.

\vspace{-6mm}
\bibliography{References}

\begin{thebibliography}{13}
\expandafter\ifx\csname natexlab\endcsname\relax\def\natexlab#1{#1}\fi

\bibitem[{Johnson-Laird and Byrne(2002)}]{johnson2002conditionals}
Philip~N Johnson-Laird and Ruth~MJ Byrne. 2002.
\newblock Conditionals: a theory of meaning, pragmatics, and inference.
\newblock \emph{Psychological review}, 109(4):646.

\bibitem[{Krishnaswamy(2017)}]{krishnaswamy2017montecarlo}
Nikhil Krishnaswamy. 2017.
\newblock \emph{Monte-Carlo Simulation Generation Through Operationalization of
  Spatial Primitives}.
\newblock Ph.D. thesis, Brandeis University.

\bibitem[{Krishnaswamy and Pustejovsky(2016)}]{krishnaswamy2016voxsim}
Nikhil Krishnaswamy and James Pustejovsky. 2016.
\newblock Vox{S}im: A visual platform for modeling motion language.
\newblock In \emph{Proceedings of COLING 2016, the 26th International
  Conference on Computational Linguistics: Technical Papers}. ACL.

\bibitem[{Li(2012)}]{li2012hand}
Yi~Li. 2012.
\newblock Hand gesture recognition using kinect.
\newblock In \emph{2012 IEEE International Conference on Computer Science and
  Automation Engineering}, pages 196--199. IEEE.

\bibitem[{Matuszek et~al.(2014)Matuszek, Bo, Zettlemoyer, and
  Fox}]{matuszek2014learning}
Cynthia Matuszek, Liefeng Bo, Luke Zettlemoyer, and Dieter Fox. 2014.
\newblock Learning from unscripted deictic gesture and language for human-robot
  interactions.
\newblock In \emph{AAAI}, pages 2556--2563.

\bibitem[{McNeely-White et~al.(2019)McNeely-White, Ortega, Beveridge, Draper,
  Bangar, Patil, Pustejovsky, Krishnaswamy, Rim, Ruiz et~al.}]{mcneely2019user}
David~G McNeely-White, Francisco~R Ortega, J~Ross Beveridge, Bruce~A Draper,
  Rahul Bangar, Dhruva Patil, James Pustejovsky, Nikhil Krishnaswamy, Kyeongmin
  Rim, Jaime Ruiz, et~al. 2019.
\newblock User-aware shared perception for embodied agents.
\newblock In \emph{2019 IEEE International Conference on Humanized Computing
  and Communication (HCC)}, pages 46--51. IEEE.

\bibitem[{Pustejovsky and Krishnaswamy(2016)}]{pustejovsky2016LREC}
James Pustejovsky and Nikhil Krishnaswamy. 2016.
\newblock {VoxML}: A visualization modeling language.
\newblock \emph{Proceedings of LREC}.

\bibitem[{Pustejovsky and Krishnaswamy(2019)}]{pustejovsky2019situational}
James Pustejovsky and Nikhil Krishnaswamy. 2019.
\newblock Situational grounding within multimodal simulations.
\newblock \emph{arXiv preprint arXiv:1902.01886}.

\bibitem[{Roque et~al.(2020)Roque, Tsuetaki, Sarathy, and
  Scheutz}]{roque2020developing}
Antonio Roque, Alexander Tsuetaki, Vasanth Sarathy, and Matthias Scheutz. 2020.
\newblock Developing a corpus of indirect speech act schemas.
\newblock In \emph{Proceedings of The 12th Language Resources and Evaluation
  Conference}, pages 220--228.

\bibitem[{Tavares and Padilha(1995)}]{tavares1995new}
Jo{\~a}o Manuel Ribeiro~Silva Tavares and Armando Jorge Monteiro~Neves Padilha.
  1995.
\newblock A new approach for merging edge line segments.
\newblock \emph{Proceedings RecPad'95, Aveiro}.

\bibitem[{Tellex et~al.(2020)Tellex, Gopalan, Kress-Gazit, and
  Matuszek}]{tellex2020robots}
Stefanie Tellex, Nakul Gopalan, Hadas Kress-Gazit, and Cynthia Matuszek. 2020.
\newblock Robots that use language.
\newblock \emph{Annual Review of Control, Robotics, and Autonomous Systems},
  3:25--55.

\bibitem[{Tzafestas and Tzafestas(1991)}]{tzafestas1991blackboard}
Spyros Tzafestas and Elpida Tzafestas. 1991.
\newblock The blackboard architecture in knowledge-based robotic systems.
\newblock In \emph{Expert systems and robotics}, pages 285--317. Springer.

\bibitem[{Williams et~al.(2019)Williams, Bussing, Cabrol, Boyle, and
  Tran}]{williams2019mixed}
Tom Williams, Matthew Bussing, Sebastian Cabrol, Elizabeth Boyle, and Nhan
  Tran. 2019.
\newblock Mixed reality deictic gesture for multi-modal robot communication.
\newblock In \emph{2019 14th ACM/IEEE International Conference on Human-Robot
  Interaction (HRI)}, pages 191--201. IEEE.

\end{thebibliography}
\bibliographystyle{acl_natbib}

\end{document}